\documentclass[letterpaper, 10 pt, conference]{ieeeconf}  

\IEEEoverridecommandlockouts                              

\usepackage{cite}
\usepackage{amsmath,amssymb,amsfonts}
\usepackage{algorithmic}
\usepackage{graphicx}
\usepackage{textcomp}
\usepackage{xcolor}
\usepackage{hyperref}
\usepackage{eurosym}
\usepackage{array}
\usepackage{makecell}
\usepackage{tikz}%
\usepackage{pgfplots}%
\usepgfplotslibrary{external}%
\pgfplotsset{compat=1.15}
\usepackage{algorithm}%
\usepackage{algorithmic}%
\usepackage{verbatim}%
\usepackage{float}%

\usepackage{multirow}
\usepackage{colortbl}
\usepackage{booktabs}
\usepackage{hhline}

\hypersetup{
    colorlinks=true,
    linkcolor=blue,
    urlcolor=blue
    }
\usepackage[top=25.4mm,bottom=19.1mm,right=19.1mm,left=19.1mm]{geometry}
\def\BibTeX{{\rm B\kern-.05em{\sc i\kern-.025em b}\kern-.08em
    T\kern-.1667em\lower.7ex\hbox{E}\kern-.125emX}}

\begin{document}

\title{\LARGE \bf
Towards Remote Robotic Competitions: \\ An Internet-Connected Task Board and Dashboard
}

\author{Peter So$^{*}$, Jonas Wittmann$^{\dagger}$, Patrick Ruhkamp$^{\ddagger}$, Andriy Sarabakha$^{*}$and Sami Haddadin$^{*}$
\thanks{$^{*}$Peter So, Andriy Sarabakha, and Sami Haddadin are with the Munich Institute of Robotics and Machine Intelligence, TU Munich,
        Germany
        {\tt\small peter.so@tum.de, andriy.sarabakha@tum.de, haddadin@tum.de}}%
\thanks{$^{\dagger}$Jonas Wittmann is with the Chair of Applied Mechanics, School of Engineering \& Design, Munich Institute of Robotics and Machine Intelligence, TU Munich, Germany
        {\tt\small jonas.wittmann@tum.de}}%
\thanks{$^{\ddagger}$Patrick Ruhkamp is with the Chair of Computer Assisted Medical Procedures and Augmented Reality, TU Munich,
        Germany
        {\tt\small p.ruhkamp@tum.de}}%
}

\maketitle
\thispagestyle{empty}
\pagestyle{empty}

\begin{abstract}
In this work we present a platform to assess robot platform skills using an internet-of-things (IoT) task board device to aggregate performances across remote sites. We demonstrate a concept for a modular, scale-able device and web dashboard enabling remote competitions as an alternative to in-person robot competitions. We share data from nine robot platforms located across four continents in three manipulation task categories of \textit{object localization}, \textit{object insertion}, and \textit{component disassembly} through an organized international robot competition -- the \textit{Robothon Grand Challenge}. This paper discusses the design of an electronic task board, the strategies implemented by the top-performing teams and compares their results with a benchmark solution to the presented task board. Through this platform, we demonstrate fully remote, online competitions can generate innovative robotic solutions and tested a tool for measuring remote performances. Using the open-sourced task board code and design files, the reader can reproduce the benchmark solution or configure the platform for their own use case and share their results transparently without transporting their robot platform.
\end{abstract}

\section{Introduction}
\begin{figure}[!t]
\centering
\includegraphics[width=0.95\columnwidth]{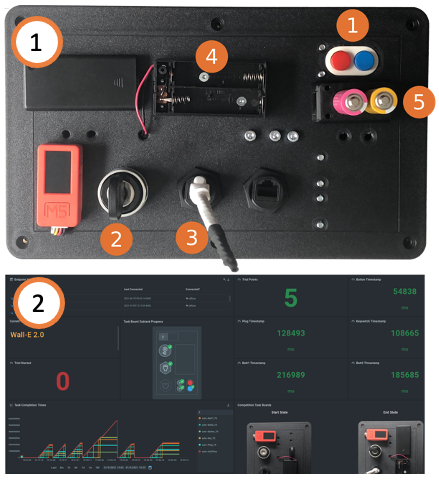}
\caption{Developed IoT task board and web dashboard used for the competition monitoring. The task board integrates five tasks of the categories \textit{object localization}, \textit{object insertion} and \textit{component disassembly} with increasing difficulty: Button Push (1), Key Switch (2), Ethernet Port (3), Battery Removal (4), Battery Recycling (5). An internet dashboard tracks each task board and allows performance data to be remotely viewed in near real-time (5 second publishing rate) and historical data downloaded by the public.}
\label{fig:IoTTaskBoardAndDashboard}
\end{figure}

Competitions play an important role in developing the field of robotics by providing an arena for exhibiting state-of-the-art methods on a constrained problem and time frame \cite{Competitions_for_Benchmarking_2015}.
However, organizing challenges involves considerable effort from both the organizers and participants to arrive at a meaningful outcome. Traditionally, robot competitions have the logistical burden of transporting equipment across state borders resulting in shipping costs, lost parts, and stress-inducing interactions with customs agents to bring a robot platform to participate in-person at conferences \cite{AWSPickingChallenge2016}. Organizers need to set up a comprehensive set of rules for judging committees to ensure a fair competition which often involves in-person inspections and witnessing of performances to assert the validity of trial attempts \cite{SubT}. Building benchmarks around common household objects, like \cite{calli_walsman_singh_srinivasa_abbeel_dollar_2015}, is an imperfect solution as manufacturers change their products over time.  
Physical task boards, like the \textit{NIST} task board \cite{NIST_Assembly_Benchmarks}, provide a system to compare robot platforms across categories of manipulation skills but still rely on expert witnessed performances and self-assembly of kits. Completely simulated competitions offer the convenience of digital grading, portability, and easily reset environments, however simulated robot solutions are criticized for their lack of robustness and failure to transfer to real-world scenarios \cite{OCRTOC_2021} \cite{james2019rlbench}. Human manipulation performance is measured with standardized test kits like the \textit{Box and Block Test} \cite{desrosiers_bravo_hebert_dutil_mercier_1994} and \textit{Minnesota Manual Dexterity Test} \cite{desrosiers_rochette_hebert_bravo_1997_minnesota} to assess dexterity and coordination across several participants. Similarly, we present a novel assessment system comprised of a physical task board with built-in grading and reporting and corresponding web dashboard to provide a transparent and objective comparison tool for measuring manipulation performance for versatile robotic tasks across multiple robot platforms shown in Fig.~\ref{fig:IoTTaskBoardAndDashboard}. \newline

We developed a portable, desktop competition platform which uses electrical circuits to confirm task completion and an internet-of-things (IoT) device to automatically publish trial results to a publicly accessible website to provide insight into remote performances. A lightweight assessment platform eliminates the logistical burden and cost of transporting entire robot platforms to a central location, and instead can be shipped directly to participants, or assembled by themselves with low-cost parts, to conduct experiments in their own lab. This enables participants from distant locations to participate in international competitions and opens the opportunity for continuous, asynchronous demonstrations to be recorded and incremental performance improvements to be duly recognized. Besides the main contribution of this paper to present the IoT developed task board, we also present data from a field study on robot manipulation tasks relevant to e-waste handling and recycling grouped into three skill categories: \textit{object localization}, \textit{object insertion}, and \textit{component disassembly}. Additionally, we provide an exemplary solution to the presented task board using a vision guided, force sensitive robot to encourage others to reproduce and improve on the results from one of the competition teams of the \textit{Robothon Grand Challenge}.

The following sections are organized as follows. In Sec.~\ref{sec:RAT}, we present the construction of the competition task board and the web dashboard design. In Sec.~\ref{sec:Platform}, we present a benchmark solution to the task board with implementation details. Sec.~\ref{sec:results} shows competition data collected with the IoT task board for robot platforms alongside a human expert performance as the upper performance limit. The benefits of the developed task board on robotic competitions are discussed in Sec.~\ref{sec:Discussion}. Conclusions follow in Sec.~\ref{sec:Conclusion}.

\section{Remote Assessment Tools}\label{sec:RAT}
\begin{figure}[!t]
\centering
\includegraphics[width=0.95\columnwidth]{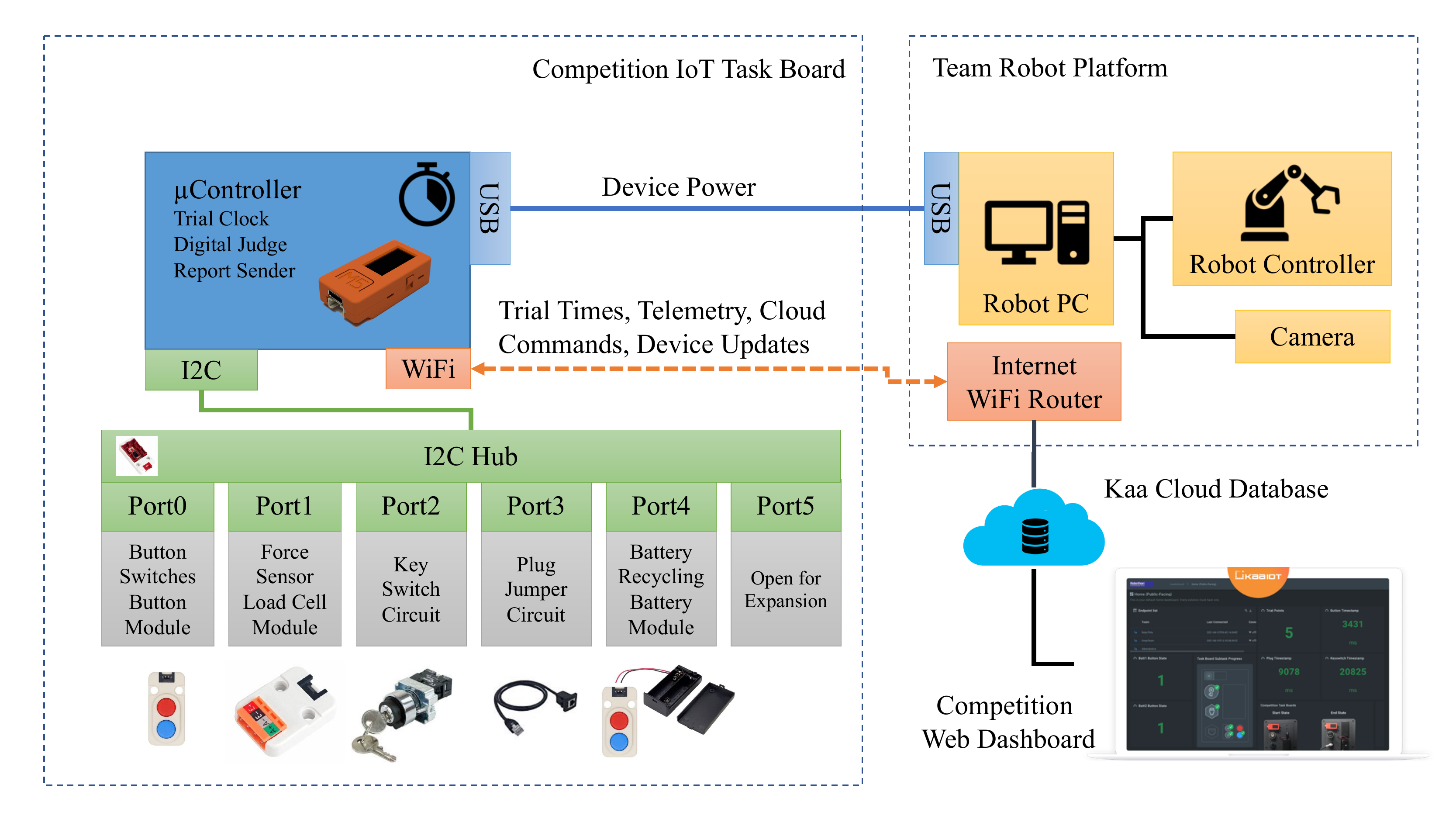}
\caption{Internet-of-things task board system architecture and connection diagram with a robot platform under test and web dashboard. The on-board microcontroller enforces competition rules and tracks task completion in miliseconds and automatically publishes results to a web dashboard.}
\label{fig:systemArchitecture}
\end{figure}

To quantitatively assess manipulation performance across several robot platforms, the task board shown in Fig.~\ref{fig:IoTTaskBoardAndDashboard} includes a built-in timer and electrical circuits connected to components mounted on the task board surface to automatically record execution times for each task. For the \textit{Robothon Grand Challenge}, components commonly found in e-waste processing facilities were selected to challenge teams to demonstrate manipulation capabilities of their robot platforms. The selected components and the circuit design shown in Fig. \ref{fig:systemArchitecture} record task completion times to approximate a robot platform's ability to locate the task board (Velcro strips and button modules), perform tight tolerance insertions (key switch and Ethernet plug), and disassemble components (battery case).

\begin{figure}[!t]
\centering
\includegraphics[width=0.95\columnwidth]{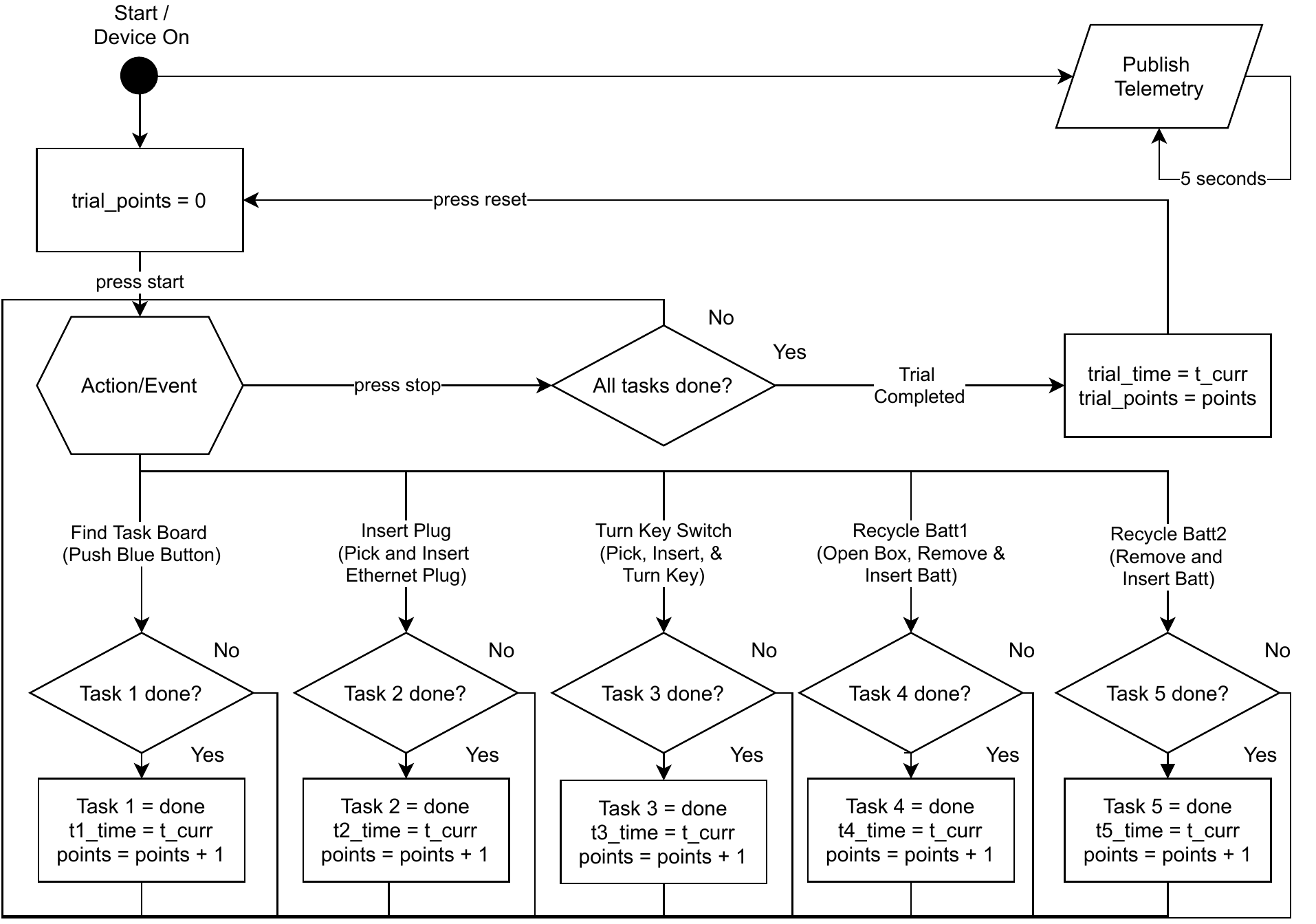}
\caption{Designed task board execution protocol and task time measurements. Each successfully completed task was confirmed by the on-board microcontroller monitoring an electrical circuit and timestamp recorded. Teams were permitted to complete tasks in their preferred order. Individual task times are computed by the different between successive timestamps. }
\label{fig taskBoardProtocol}
\end{figure}

A microcontroller enforces the experiment protocol and automatically publishes device telemetry to a web dashboard. Users connect the device to the internet over a WiFi radio by supplying their network credentials over a simple configuration website on the device. Once connected, the device reports telemetry every five seconds to a remote database and a cloud hosted server renders the data on a publicly accessible website. Task board telemetry includes individual task completion timestamps for the last active trial, currently accumulated trial points, and a summation of accelerometer readings during the last active trial to estimate the amount of device interaction. To begin a trial, the user must set the components into the starting positions, reset the task board with the Velcro strips in front of the robot platform, and press the start trial button. A program on the microcontroller requires that the task board is in the proper initial state prior to arming the start trial button which enforces uniform start conditions across multiple trials.  Teams in the competition were asked to randomly place the task board on Velcro strips on a table in front of their robot platform and complete all tasks autonomously as fast as possible with their robot platform. The corresponding protocol is shown in Fig. \ref{fig taskBoardProtocol}. A trial ends when the user completes each of the five manipulation tasks and presses the stop trial button or a 10-minute trial timer expires. All trial performance data is automatically reported and displayed on the task board's LED screen. 

The task boards are constructed from an ABS project box measuring 300~mm~x~150~mm~x~80~mm. All manipulation components are mounted onto a CNC-cut lid attached to the housing with a 3D-printed mount and suspended on a 10~kg load cell to capture the surface interaction forces. 
The \textit{M5StickC} unit by \textit{M5Stack} was selected as the on-board microprocessor for its integrated accelerometers, WiFi radio, LED screen, battery, buttons, I2C bus, and extendable low-cost sensor options. 
A \textit{PbHub}, an I2C hub device, expands the number of digital inputs so that each mounted component can have a dedicated monitoring circuit see Fig. \ref{fig:systemArchitecture}. The task board can be customized for different manipulation tasks. For the \textit{Robothon Grand Challenge} two \textit{Button modules} define the \textit{Find Task Board}, \textit{Recycle Battery 1/2}, and \textit{Stop Trial Button} tasks. 
A 22 mm panel mount key switch wired into the \textit{PbHub} and 3D-printed key holster compose the \textit{Turn Key Switch} task. 
A trimmed and jumpered Ethernet cable plug and two panel mount RJ-45 ports wired into the \textit{PbHub} comprise the \textit{Insert Plug} task. 
A dual keystone battery pack with removable lid fixed onto the task board and 3D-printed button access cover represent the \textit{Remove Battery} task. 
The entire task board assembly is designed with \textit{Onshape} and fabricated at the \textit{Munich Institute of Robotics and Machine Intelligence}. 
The microcontroller is programmed using \textit{Microsoft Visual Studio Code} and interfaced with \textit{KaaIoT} to provide remote data logging and to create a web dashboard view of trial data for each task board \cite{KaaIoT}. The task board design was tested with an in-house robot platform presented as the benchmark solution in Sec. \ref{sec:Platform}

Each task board is assigned a unique endpoint ID to distinguish each team's data on a competition web dashboard. Telemetry from all task boards are displayed on the competition website to show usage statistics and to stimulate competition between teams during the competition. 
Team solutions were evaluated with a live demonstration over video conference with the remote jury who referenced the dashboard to confirm completion of the manipulation tasks. 
Data from the competition and the web dashboard can be viewed at \href{https://bit.ly/robothonDashboard}{https://bit.ly/robothonDashboard}. 

\section{A Benchmark Solution to the Task Board}\label{sec:Platform}

\begin{table}[!hb]
    \centering
    \caption{Teams' robot platforms used for the competition.}
    \resizebox{\columnwidth}{!}{
    \setlength\tabcolsep{3pt}
    \begin{tabular}{l|cccccccc|cccccccc|ccc|cccc|c}
    
        \multicolumn{1}{c|}{\centering Robothon} & \multicolumn{8}{c|}{\multirow{2}{*}{Arm Manipulator}} & 
        \multicolumn{8}{c|}{\multirow{2}{*}{End Effector}} & 
        \multicolumn{7}{c|}{Sensor} & \\
        \multicolumn{1}{c|}{\centering 2021}& &&&&&&&& &&&&&&&& \multicolumn{3}{c}{Vision} & \multicolumn{4}{c|}{Force} &\\
        \hline
        {\rotatebox[origin=c]{90}{\parbox[t][][c]{3.5cm}{\centering Robot Components}}} & 
        {\rotatebox[origin=c]{90}{\parbox[t][][c]{3.5cm}{\centering Kuka LBR iiwa}}} & 
        {\rotatebox[origin=c]{90}{\parbox[t][][c]{3.5cm}{\centering Franka Emika Arm}}} & 
        {\rotatebox[origin=c]{90}{\parbox[t][][c]{3.5cm}{\centering Universal Robots UR10e}}} & 
        {\rotatebox[origin=c]{90}{\parbox[t][][c]{3.5cm}{\centering Universal Robots UR5}}} &
        {\rotatebox[origin=c]{90}{\parbox[t][][c]{3.5cm}{\centering Universal Robots UR5e}}} & 
        {\rotatebox[origin=c]{90}{\parbox[t][][c]{3.5cm}{\centering ABB IRB120}}} & 
        {\rotatebox[origin=c]{90}{\parbox[t][][c]{3.5cm}{\centering Dobot Magician}}} & 
        {\rotatebox[origin=c]{90}{\parbox[t][][c]{3.5cm}{\centering Igus RL-D BRT}}} &
        
        {\rotatebox[origin=c]{90}{\parbox[t][][c]{3.5cm}{\centering Robotiq Hand-E}}} &
        {\rotatebox[origin=c]{90}{\parbox[t][][c]{3.5cm}{\centering Schunk WSG 50}}} &
        {\rotatebox[origin=c]{90}{\parbox[t][][c]{3.5cm}{\centering Festo HGP-16-A-B-SSK}}} &
        {\rotatebox[origin=c]{90}{\parbox[t][][c]{3.5cm}{\centering Franka Emika Hand}}} &
        {\rotatebox[origin=c]{90}{\parbox[t][][c]{3.5cm}{\centering 20W Electromagnet}}} &
        {\rotatebox[origin=c]{90}{\parbox[t][][c]{3.5cm}{\centering Dobot Magician Gripper}}} &
        {\rotatebox[origin=c]{90}{\parbox[t][][c]{3.5cm}{\centering Dynamixel CL430}}} &
        {\rotatebox[origin=c]{90}{\parbox[t][][c]{3.5cm}{\centering 3D Printed Fingers}}} &
        
        {\rotatebox[origin=c]{90}{\parbox[t][][c]{3.5cm}{\centering On-Arm Intel RS-D435}}} &
        {\rotatebox[origin=c]{90}{\parbox[t][][c]{3.5cm}{\centering On-Arm Intel RS-D435i}}} &
        {\rotatebox[origin=c]{90}{\parbox[t][][c]{3.5cm}{\centering Fixed Microsoft Azure}}} &
        
        {\rotatebox[origin=c]{90}{\parbox[t][][c]{3.5cm}{\centering OnRobot Hex}}} &
        {\rotatebox[origin=c]{90}{\parbox[t][][c]{3.5cm}{\centering Robotiq FT 300-S}}} &
        {\rotatebox[origin=c]{90}{\parbox[t][][c]{3.5cm}{\centering ATI F/T Gamma}}} &
        {\rotatebox[origin=c]{90}{\parbox[t][][c]{3.5cm}{\centering Built-in Arm Torque}}} &
        
        {\rotatebox[origin=c]{90}{\parbox[t][][c]{3.5cm}{\centering Component Count}}}\\
        \hline
        RoboTHIx    & 1 & - & - & - & - & - & - & -     & - & 1 & - & - & 1 & - & - & -     & - & - & -   & - & - & - & 1 & 4\\
        RoboPIG     & - & - & - & 1 & - & - & - & -     & 1 & - & - & - & - & - & - & 1     & 1 & - & -   & - & - & - & 1 & 5\\
        Ewas        & - & - & 1 & - & - & - & - & -     & 1 & - & - & - & - & - & - & 1     & - & - & 1   & - & 1 & - & - & 5\\
        RAND-E      & - & - & - & - & - & 1 & - & -     & - & - & 1 & - & - & - & - & 1     & 1 & - & -   & - & - & 1 & - & 5\\
        WallE2.0   & - & 1 & - & - & - & - & - & -     & - & - & - & 1 & - & - & - & 1     & - & 1 & -   & - & - & - & 1 & 5\\
        ECODOBO     & - & - & - & - & - & - & 1 & -     & - & - & - & - & - & 1 & - & -     & - & 1 & -   & - & - & - & - & 3\\
        Schmalkalden& - & - & - & - & - & - & - & 1     & - & - & 1 & - & - & - & - & 1     & 1 & - & -   & - & - & - & - & 4\\
        Stevens'    & - & - & - & - & 1 & - & - & -     & - & - & - & - & - & - & 1 & 1     & 1 & - & -   & 1 & - & - & - & 5\\
        Pandaria One& - & 1 & - & - & - & - & - & -     & - & - & - & 1 & - & - & - & 1     & - & - & -   & - & - & - & 1 & 4\\
        \hline
        \textbf{SUM}
                    &\textbf{1}&\textbf{2}&\textbf{1}&\textbf{1}&\textbf{1}&\textbf{1}&\textbf{1}&\textbf{1}&\textbf{2}&\textbf{1}&\textbf{2}&\textbf{2}&\textbf{1}&\textbf{1}&\textbf{1}&\textbf{7}&\textbf{4}&\textbf{2}&\textbf{1}&\textbf{1}&\textbf{1}&\textbf{1}&\textbf{4}&\textbf{40}\\
        
    \end{tabular}
    }
    \label{tab:RobotPlatformsSideBySide}
\end{table}

\begin{figure}[t]
    \centering
    \includegraphics[width=0.95\columnwidth]{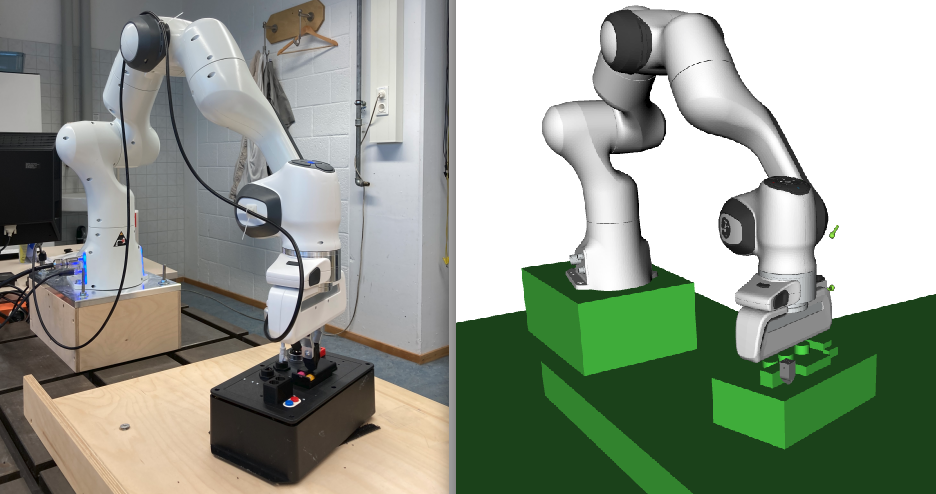}
    \caption{Hardware setup of the benchmark solution with the implemented environment model for computing feasible motion plans.}
    \label{fig:BenchmarkSolution}
\end{figure}

A major benefit of the IoT task board is its versatility in measuring the performance of several different robot platforms. During the \textit{Robothon Grand Challenge}, a diverse set of robot platforms was evaluated as shown in Table \ref{tab:RobotPlatformsSideBySide}. We present an example solution to the presented challenge depicted in Fig.~\ref{fig:BenchmarkSolution} to provide a benchmark for the robotics community to reproduce performance results and improve upon with their own performance upgrades. The benchmark solution offers a starting point for advanced manipulation skill development with a vision-guided, force-sensitive seven degrees of freedom robot arm. 
An arm-mounted \textit{Intel RealSense D435i} camera detects the initial location of the task board and integrated joint torque sensors provide tactile feedback for delicate manipulation tasks.
Rubber coated, 3D-printed fingers fixed to the \textit{FRANKA EMIKA} electric pinch gripper provide the required dexterity to complete all the tasks on the task board. 
Four software modules are implemented for the complete execution process. A high-level state machine controls the overall task execution and has implemented error detection as well as basic error recovery strategies. Via service calls, the state machine triggers the implemented features in the other modules, i.e. vision, motion planner and task control. The communication between components is implemented in \textit{ROS} \cite{ROS} \textit{Melodic} and the software modules run on a real-time capable robot control computer with 64-bit Ubuntu 18.04. 
The implementation of the benchmark robot solution is explained in next two sections. 

\subsection{Task Board Localization}
The benchmark solution solves the tasks based on a reference frame that is attached to the corner of the task board. To successfully interact with the task board components, the robot moves above the task board area and acquires an overhead image then calculates the task board reference frame with respect to its end effector frame. A hand-eye calibration~\cite{hand_eye_calib} is performed to compute the static transformation between the camera frame and the robot's end effector frame. 
A classical pose estimation approach ~\cite{10.1007/978-3-030-71051-4_10} first detects robust features on the task board and matches them with a pre-acquired target image of the object. Then, a homographic transformation between the current view and the target is computed with a robust \textit{RANSAC} scheme to filter out noisy matches. Directional vectors on the surface of the task board are computed and three co-planar points in 3D are acquired by leveraging the depth information from the active depth sensor. These three points define the planar surface of the task board lid regarding its center point and known dimensions and hence its 6D pose, which is published as pose array for the subsequent robotic manipulation tasks. 
The pose of the task board is defined in the reference frame of the vision system.
The benchmark solution achieves a mean pose estimation error of $0.98\pm 0.35\text{cm}$ for translation and $0.72\pm0.32^{\circ}$ for rotation across eight trials for the vision-based approach against the ground-truth pose of the task board, which is determined by measuring the 3D coordinates of four corner points of the task board through the forward kinematics of the robot's end effector. 

\subsection{Robotic Manipulation}
Similar to \cite{Dai2016}, we define a \textit{Parameterizable Compliant Motion Controller (PCMC)} for the task control module to solve all five tasks by adapting its parameters. The \textit{PCMC} is based on a hybrid position/force control scheme with three parameterizable inputs. For tactile interaction tasks like the button a desired push force of the end effector can be specified. The desired Cartesian compliance defined by stiffness and damping matrices in  $\mathbb{R}^{6\times6}$ is used to compensate position inaccuracies for lower precision tasks like plugging in the Ethernet cable. For high precision tasks, i.e. for the key insertion, a solely compliant-based approach is not sufficient. Therefore, the user can specify a desired amplitude and frequency of a spiral search motion that adapts the end effector's pose for solving the Peg-in-Hole problem. The latter is based on \cite{FrankaPlugInController}. The \textit{PCMC} is integrated in the ROS architecture using \textit{ros\_control} \cite{ROScontrol}. Due to the limited amount of time, we focused on the low-level control schemes for the task executions and therefore use the open source planning framework \textit{MoveIt!} \cite{MoveIt} to plan the motions in between. To ensure collision-free motions, we created a simplified geometry model of the environment as shown in Fig.~\ref{fig:BenchmarkSolution} and use a \textit{RRTconnect} in our motion planner module. 

\label{subsec:Tactile}
A task success is confirmed by the robot with either a reached z-position or a measured external interaction force at the end effector tip. E.g. for the push blue button task, the \textit{PCMC} is parameterized with a desired force of $F_{ref,BB} = 2.0\text{ N}$, while for the red button we set $F_{ref, RB} = 3.0\text{ N}$. The state machine triggers the \textit{PCMC} and the current estimated external force on the end effector is monitored until it reaches the desired push force. Fig.~\ref{fig::EEforces} shows the controlled end effector force $F$ over time. For both the blue task and the red button task, the task execution is stopped when the desired force is applied which is marked by the red circles. Note that our benchmark solution is not parameterized properly: The robot keeps on pushing the button although the microcontroller integrated in the task board already reports a successful push indicated by the vertical black lines. A minor push force would be sufficient.

\section{Results}\label{sec:results}
\begin{figure}[!t]
\centering
\includegraphics[width=0.95\columnwidth]{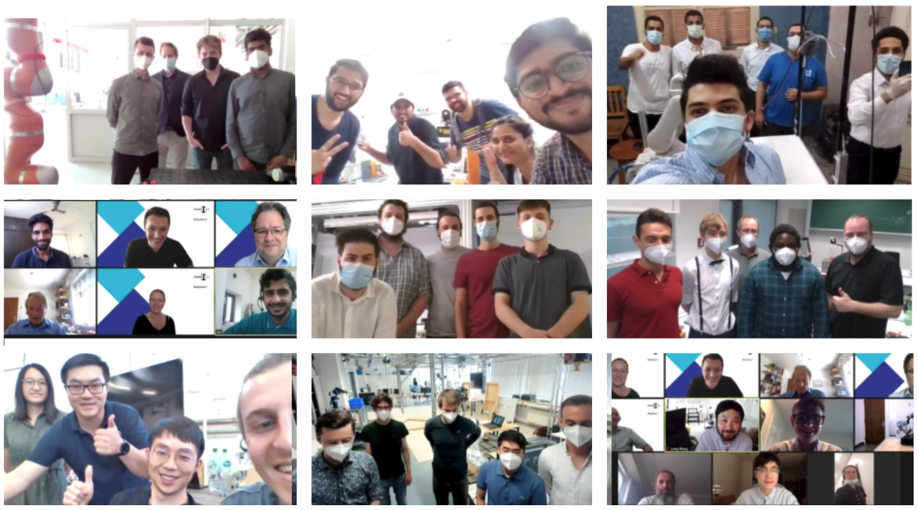}
\caption{Robothon 2021 competition teams representing nine unique robot platforms from across four continents were connected via the IoT task board as they completed their automated solutions over the 4-week development period in a completely remote robot competition. Final presentations were conducted with the competition organizers over video conference.}
\label{fig RobothonTeams}
\end{figure}

\begin{figure}[!t]
\centerline{\includegraphics[width=0.95\columnwidth]{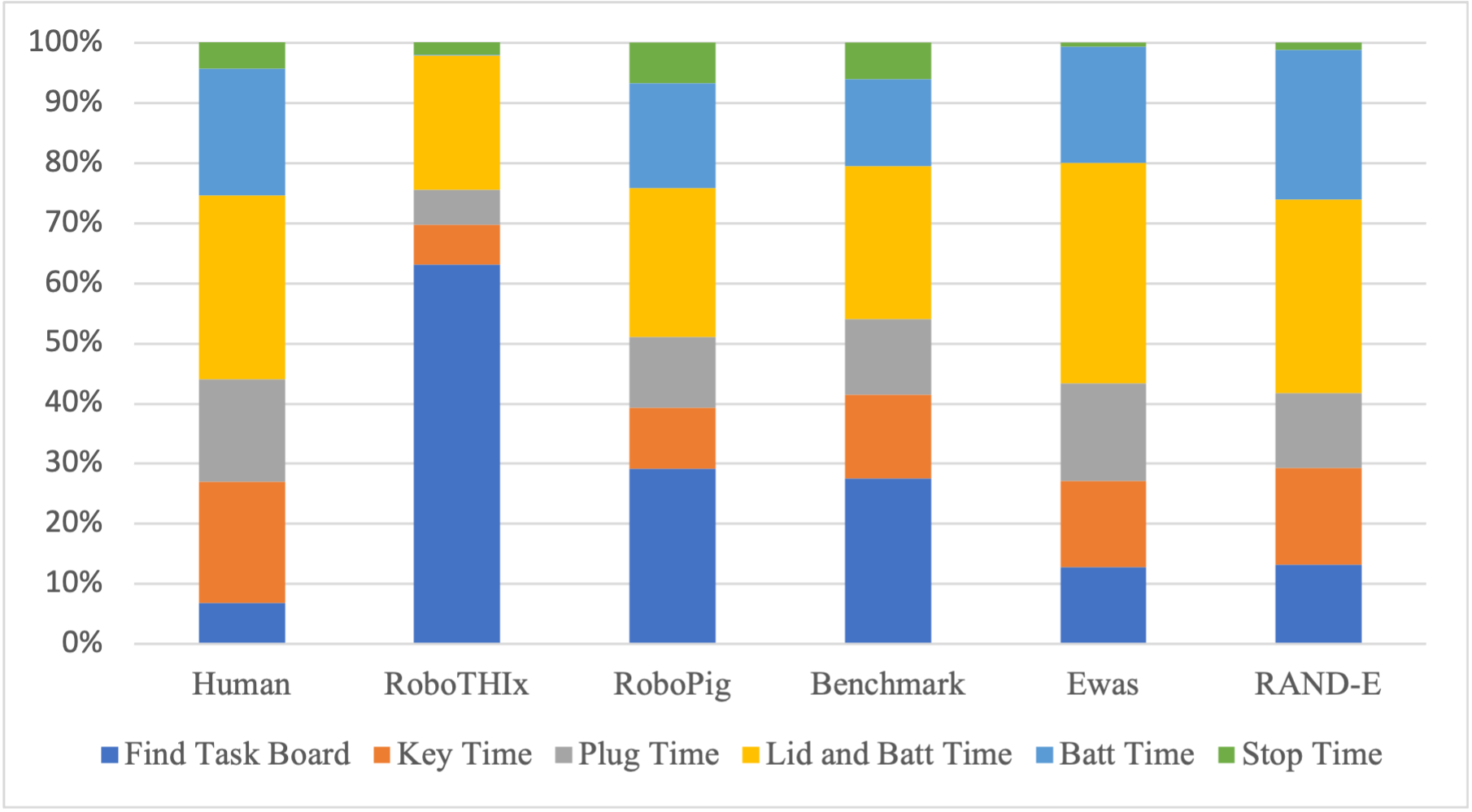}}
\caption{Task execution time as percentage of total trial time for the top four teams compared to a human demonstration. The task board shows where each actor spent their time for comparison within tasks as well as overall trial time. Team RoboTHIx spent more than half their trial time to accurately locate the task board and then swiftly completed the remaining tasks to beat the other robot teams.}
\label{fig StackedPercentageTaskTime}
\end{figure}

The competition drew 14 robot team applications of which nine teams presented final solutions with the designed task board, Fig. \ref{fig RobothonTeams}. Qualified teams were shipped a competition task board ahead of a 4-week development period to adapt and program a single arme robot platform to autonomously complete the challenge tasks. Four of the nine teams who received a task board successfully completed all tasks. Tab.~\ref{tab:TrialCompletionTimes} shows task execution times recorded by the task board for each team with the fastest times by task shown in bold. Fig. \ref{fig StackedPercentageTaskTime} shows a individual task execution time as a percentage of the entire trial time to illustrate different team strategies and where speed improvements can be realized. In addition to the results of the robot platforms, the fastest recorded time of a human completing the task board protocol with a single arm is included for reference. The next subsections discuss the different strategies employed for each task in detail.

\begin{table}[t]
    \centering
    \caption{Task completion times in seconds reported by task board.}
    \resizebox{\columnwidth}{!}{
    \begin{tabular}{l|c|c|c|c|c|c|c}
        Subject under Test & {\rotatebox[origin=c]{90}{\parbox[t][][c]{1.5cm}{\centering Find Task Board}}} & {\rotatebox[origin=c]{90}{\parbox[t][][c]{1.5cm}{\centering Activate Key Switch}}} & {\rotatebox[origin=c]{90}{\parbox[t][][c]{1.5cm}{\centering Move Plug}}} & {\rotatebox[origin=c]{90}{\parbox[t][][c]{1.8cm}{\centering Remove Lid and Recycle Batt1}}} & {\rotatebox[origin=c]{90}{\parbox[t][][c]{1.5cm}{\centering Recycle Batt2}}} & {\rotatebox[origin=c]{90}{\parbox[t][][c]{1.5cm}{\centering Press Stop Button}}} & {\rotatebox[origin=c]{90}{\parbox[t][][c]{1.5cm}{\centering Trial Time}}} \\
        \hline
        Human Dominant Hand & 0.55 & 1.61 & 1.36 & 2.43 & 1.69 & 0.35 & 8.57 \\
        \hline
        RoboTHIx & 78.65 & \textbf{5.79} & \textbf{5.10} & \textbf{19.14} & \textbf{0.15} & \textbf{1.90} & \textbf{110.73} \\ 
        RoboPig & 52.40 & 17.19 & 18.02 & 47.98 & 27.82 & 15.03 & 178.02 \\ 
        Benchmark Solution & 62.78 & 31.42 & 28.72 & 57.78 & 33.02 & 13.67 & 227.38 \\ 
        Ewas & \textbf{47.66} & 53.46 & 59.90 & 136.42 & 71.75 & 2.45 & 371.63 \\ 
        RAND-E & 57.68 & 70.30 & 54.86 & 140.68 & 108.84 & 59.74 & 437.05 \\ 
        \hline
        \hline
        Avg. Robot & 59.83 & 35.63 & 33.32 & 80.40 & 48.32 & 18.56 & 264.97 \\
        \hline
        Avg. Robot - Human & 59.28 & 34.02 & 31.96 & 77.97 & 46.63 & 18.21 & 256.40 \\
        \hline
        Fastest Robot - Human & 47.11 & 4.18 & 3.74 & 16.46 & -1.54 & 1.55 & 101.80
    \end{tabular}
    }
    \label{tab:TrialCompletionTimes}
\end{table}

\subsection{Localization "Button Press" Task}
Localizing the task board was essential for teams to complete the remaining tasks since they had to randomize the placement of the task board with each new trial. This was verified by having the robot platform press down a tactile button on the corner of the task board. In the \textit{Robothon Grand Challenge}, several valid solutions emerged with different efficiencies as seen by the proportion of trial time spent on this task in Fig.~\ref{fig StackedPercentageTaskTime}. Interestingly, the fastest team, RoboTHIx, spent the most time on this task and then sped through the remaining tasks to achieve the fastest trial time. They used a tactile scanning 4-point touch method to define a work frame for the task board (three points to define a corner of the task board plane and a fourth vertical touch point to define height) while the other teams relied on a vision inspection. Team Ewas had the fastest time in this category using a fixed overhead camera to locate the task board saving robot motion time to acquire the image over teams who used arm-mounted cameras. 

\subsection{Insertion "Key Switch and Ethernet Plug" Tasks}
\label{subsec:Insertion}
\begin{figure}[!b]
	\centering
	\resizebox{0.95\columnwidth}{!}{%
		\input{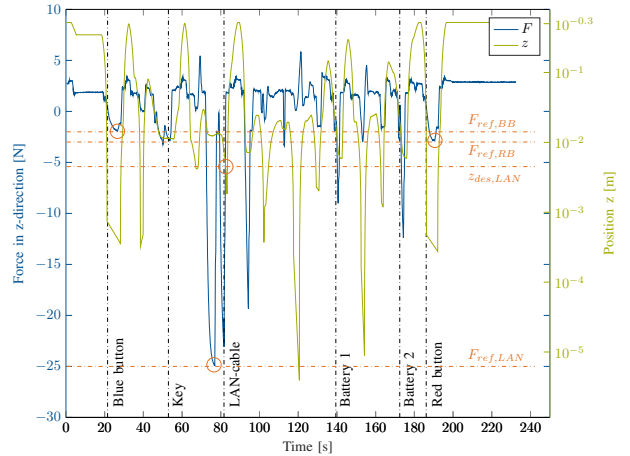}
	}
	\caption{Estimated external end effector forces and end effector position over time for one task board run. The blue line shows the interaction forces in z-direction w.r.t. the robot's base frame. The green line shows the end effector's z-position in that frame. The vertical black line show the timestamps at which the microcontroller integrated in the task board reported success for the corresponding task. The horizontal orange lines show the reference push forces $F_{ref}$ with which the \textit{PCMC} was parameterized to execute a task and the desired z-position $z_{des}$ that were used to evaluate insertion success in the state machine. The force and position values are sampled at a rate of 1.0 kHz and are smoothed out to remove noise.}
	\label{fig::EEforces}
\end{figure}

The key task involved the robot grasping a key from a holster then inserting it into a keyhole and turning the key to activate a key switch. Similarly, the plug task involved unplugging an Ethernet plug from the task board starting port and inserting it fully into the adjacent port. Teams presented similar Cartesian move behaviors for the grasp and used a version of force control, Cartesian impedance control and spiral search for the insertion movement. Team RoboTHIx recorded the fastest times, moving between predefined points in the task board work frame with tuned joint impedance values to cope with minor misalignment. 
All teams re-gripped and nudged down the Ethernet plug after their initial insertion to ensure the cable was fully seated. \newline
Wihtin the benchmarks solution, the insertion tasks are executed by the \textit{PCMC} by adding appropriate Cartesian stiffness together with the desired push force. For the Ethernet task, the \textit{PCMC} was parameterized with a reference push force of $F_{ref, LAN} = 25.0\text{ N}$  and translational Cartesian stiffness of $\mathbf{K} = \text{diag}(500.0, 500.0, 100.0)\text{ N/m}$. In contrast to the push button task the current end effector force $F$ was not used as a success criteria but instead the end effector's target z-position was set to $x_{des} = 0.0045\text{ m}$. This helps to detect errors during task execution, e.g. when the end effector gets stuck on the Ethernet port due to inaccuracies of the task board pose estimation. In case, the end effector does not reach the desired z-position, a reattempt is started. Within the task board run depicted in Fig.~\ref{fig::EEforces}, the Ethernet insertion task had to be repeated once. One clearly sees that during the first attempt, the \textit{PCMC} controls the push fore to the desired reference value of $25.0\text{ N}$. However, task success is not reported as the desired z-position could not be reached. Only in the second attempt, the end effector's z-position indicated by the green line undergoes the desired value and the state machine stops the task execution and the microcontroller reports success. A similar task control scheme was applied to the battery and key insertion with overlayed spiral search motion for the latter.

\subsection{Disassembly "Battery Removal and Recycling" Task}
Battery removal is a contact-rich, real-world task involving the opening of a sliding battery case door, prying and removing batteries from a keystone holder, and inserting the batteries into a goal battery recycling hole. All teams used the tip of their end effector to slide off the battery cover and pry out batteries from their keystones. In this task, Team RoboTHIx outperformed the human by using a high-power electromagnet to remove and recycle both batteries in one motion, resulting in the faster time in Tab. \ref{tab:TrialCompletionTimes}. \newline
Our benchmark solution approach uses the motion planner module to move the end effector between Cartesian poses that define the disassembly step. The \textit{PCMC} applies forces during an interaction motion as in sliding open the battery case door and inserting the batteries. The implemented success indicators were similar to subsection~\ref{subsec:Tactile}~and~\ref{subsec:Insertion}.

\section{Discussion}\label{sec:Discussion}
The need to develop a remote evaluation system for robot competitions arose from the uncertainty of planning a robot competition amidst the COVID19 pandemic. A portable task board provided a bankable solution for coordinating a scheduled event that could be distributed across multiple locations. A primary concern from participating teams was transporting their robot platform, auxiliary equipment, and themselves to a central location in order to participate in the robotic challenge, especially with several dynamic travel and gathering restrictions in place. To preserve the cohesiveness of a live event and to secure the integrity of the competition, the addition of a digital judge microcontroller to the task board was proposed. 
The competition was held in a completely remote format over video conferencing and the web dashboard provided near real-time telemetry of each team's progress which added a new dimension of interaction between participants and the jury members. 
The diversity of solutions impressed the jury and sparked a debate over general solutions and precise direct methods based on objective data from the task board.
The remote competition format attracted applicants from distant countries spanning multiple time zones. The asynchronous reporting and interaction log files from the IoT task board allowed organizers to track team's progress regardless of time of day. The remote concept was celebrated at the online \textit{automatica sprint} conference with over 1500 viewers watching the final presentations.  
From our experience organizing the \textit{Robothon Grand Challenge}, we see the remote robot competition enabled by IoT devices as a viable alternative to in-person competitions.

\section{Conclusion}\label{sec:Conclusion}
In conclusion, remote competitions with IoT assessment platforms offer logistical advantages over in-person events. Integrated microcontrollers on task boards can enforce competition rules and automatically publish results serve as a new objective form of collaboration between remote researchers. This work presents an IoT device, a web platform transparently monitor performances, and an example robotic solution to the task board. 
The presented platform was validated by organizing the international competition, \textit{Robothon Grand Challenge} and is planned to repeat every other year alongside the \textit{automatica} trade show. 

A video presentation of the competition results can be found at \href{http://tiny.cc/robothon2021Compilation}{http://tiny.cc/robothon2021Compilation}. Details of the competition are available at the competition website:  \href{https://www.robothon-grand-challenge.com}{www.robothon-grand-challenge.com}. 

\section*{Acknowledgments}
The development of the IoT task board was supported by Jan Harder and Maria Danninger for procurement, Burak G\"uner and Felix Spang for fabrication, Anna Hecht for assembly, Denys Buzhor for code wrangling and Barbara Schilling for event logistics. We also thank the \textit{Robothon Grand Challenge} jury for their trust and support. This work was performed at the Munich Institute of Robotics and Machine Intelligence in collaboration with industry partners Messe M{\"u}nchen and Microsoft.

\bibliographystyle{./bibliography/IEEEtran}
\bibliography{IEEEabrv,./bibliography/IEEEexample}
\end{document}